\documentclass[10pt,twocolumn,letterpaper]{article}

\usepackage{iccv}
\usepackage{times}
\usepackage{epsfig}
\usepackage{graphicx}
\usepackage{amsmath}
\usepackage{amssymb}
\usepackage{subfigure}

\usepackage[breaklinks=true,bookmarks=false]{hyperref}

\iccvfinalcopy % *** Uncomment this line for the final submission

\def\iccvPaperID{****} % *** Enter the ICCV Paper ID here

\begin{document}

%%%%%%%%% TITLE
\title{Multi-Scale Dual-Branch Fully Convolutional Network for Hand Parsing}

\author{Yang Lu\\Beihang University\\
\\
{\tt\small luyanger1799@buaa.edu.cn}
\and
Xiaohui Liang\\
Beihang University\\
\\
{\tt\small liang\_xiaohui@buaa.edu.cn}
\and
 Frederick W. B. Li\\
 University of Durham\\
 \\
 {\tt\small frederick.li@durham.ac.uk}
}

\maketitle
%\thispagestyle{empty}

%%%%%%%%% ABSTRACT
\begin{abstract}
Recently, fully convolutional neural networks (FCNs) have shown 
significant performance in image parsing, including scene parsing
and object parsing. Different from generic object parsing tasks,
hand parsing is more challenging due to small size, complex structure,
heavy self-occlusion and ambiguous texture problems. In this
paper, we propose a novel parsing framework, Multi-Scale Dual-Branch
Fully Convolutional Network (MSDB-FCN), for hand parsing tasks.
Our network employs a Dual-Branch architecture to extract features of
hand area, paying attention on the hand itself. These features are used
to generate multi-scale features with pyramid pooling strategy.
In order to better encode multi-scale features, we design a
Deconvolution and Bilinear Interpolation Block (DB-Block) for upsampling
and merging the features of different scales. To address data imbalance,
which is a common problem in many computer vision tasks as well as
hand parsing tasks, we propose a generalization of Focal Loss,
namely Multi-Class Balanced Focal Loss, to tackle data imbalance
in multi-class classification. Extensive experiments on RHD-PARSING
dataset demonstrate that our MSDB-FCN has achieved the state-of-the-art
performance for hand parsing.

\end{abstract}

%%%%%%%%% BODY TEXT
\section{Introduction}

Image parsing aims to segment an image into different semantic regions,
providing more information to better understand the content of 
an image. It can be categorized into two types: scene parsing
and object parsing. Scene parsing focuses on analyzing the composition
of the entire scene in an image, while object parsing pays more attention
on decomposing an object into different parts.

\begin{figure}[t]
\begin{center}
\subfigure[Image]{
   \includegraphics[width=0.47\linewidth]{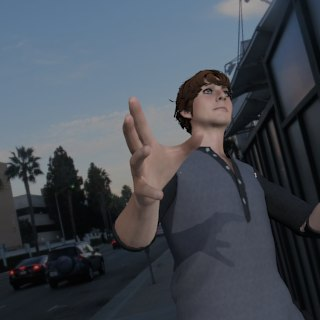}
   }
\subfigure[Ground Truth]{
   \includegraphics[width=0.47\linewidth]{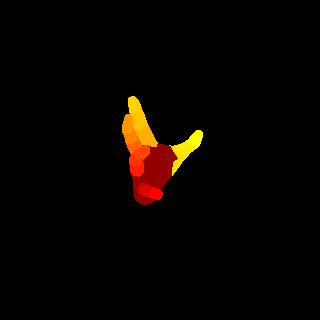}
   } 
\end{center}
   \caption{Illustration of challenges in hand parsing, including 
   complex structure, large amount of self-occlusion and ambiguous textures.
   They make hand parsing more difficult than generic object parsing tasks.}
\label{fig:cover}
\end{figure}

Recently, most researches on object parsing focus on human parsing
~\cite{DBLP:journals/corr/LiZWLLF17,DBLP:conf/cvpr/GongLZSL17,
DBLP:conf/cvpr/ZhaoLNZCWFY17,DBLP:conf/bmvc/LiAT17,DBLP:conf/cvpr/XiaWCY17}.
Compared to human parsing, hand parsing is more challenging, 
because of small size, similar textures and heavy self-occlusion of hands.
Successful hand parsing techniques have many potential applications, such
as hand behavior analysis, sign language recognition and human-machine
interaction.

State-of-the-art frameworks for image parsing are mostly based on the Fully
Convolutional Networks (FCNs)~\cite{DBLP:conf/cvpr/LongSD15}, 
which have shown excellent performance on several
benchmarks. One of the key success factors of these methods is
involving multi-scale information~\cite{DBLP:conf/cvpr/ZhaoSQWJ17,DBLP:journals/corr/ChenPSA17,DBLP:conf/cvpr/LinMSR17} 
or prior knowledge~\cite{DBLP:conf/iccv/HungTSLSL017,DBLP:conf/icmcs/HuSSS18,DBLP:conf/cvpr/FangLFXTL18}.
Multi-scale features are helpful for accurate prediction of ambiguous
pixels, by aggregating contextual information from different regions of
an image. One way to assemble multi-scale features is using the encoder-decoder
structure~\cite{DBLP:conf/cvpr/LinMSR17,DBLP:conf/cvpr/ChenWPZYS18,
DBLP:conf/eccv/ZhengJWLF18,DBLP:conf/cvpr/DingJSL018,DBLP:conf/eccv/LinJLCH18}. 
After generating the features, some strategies can be used to encode them
together. Another approach is using multi-scale pooling strategies~\cite{DBLP:conf/eccv/HeZR014,
DBLP:conf/cvpr/ZhaoSQWJ17,DBLP:conf/aaai/YangLCGCL18,DBLP:conf/eccv/WangXQZZY18}, such
as pyramid pooling. By controlling the pool size of average- or max-pooling, 
multi-scale features can be obtained.
Similarly, embedding some semantic prior knowledge as a guide can also improve
the performance of image parsing~\cite{DBLP:conf/iccv/HungTSLSL017,
DBLP:conf/icmcs/HuSSS18,DBLP:conf/cvpr/FangLFXTL18,DBLP:conf/cvpr/XiaWCY17}. 
Some studies~\cite{DBLP:conf/icmcs/HuSSS18,DBLP:conf/cvpr/FangLFXTL18} have demonstrated that results can be improved by feeding a coarse-grained result
to networks, providing effective contextual clues for a fine-grained
parsing.

In this paper, we propose a framework for hand parsing. Different
from scene parsing, it is similar to an object parsing task paying more
attention to an object itself rather than the category of a scene. Because
some background information may become a disturbance to accurate parsing.
In contrast to human body, a hand is more complex in structure,
smaller in size, comprising many similar textures and heavy in self-occlusion.
These properties make hand parsing more challenging.

We therefore design a novel Multi-Scale Dual-Branch Fully Convolutional Network
(MSDB-FCN) comprising a parsing branch and a mask branch to better extract 
multi-scale features of a hand region. The role of the mask branch is to
calculate hand position and provide prior information for the parsing
branch. In the parsing branch, by extracting features and merging with
the prior information of the hand area, the negative impact of background
is reduced, which improves the parsing accuracy. Moreover, considering
the small size and complex structure of a hand, our MSDB-FCN employs more reasonable
pool sizes by using multi-scale pooling to obtain multi-scale features, 
instead of using the original pyramid pooling strategy. For
multi-scale features fusion, we design a new upsampling and context coding structure,
namely Deconvolution and Bilinear Interpolation Block (DB-Block), to
encode multi-context information of the hand. Compared with using either deconvolution
or bilinear interpolation alone for upsampling, the model of using DB-Block gets 
better results.

%%updated
On the other hand, the problem of data imbalance between categories 
significantly magnifies due to the small hand size. In response to such a
situation, we propose a Multi-Class Balanced Focal Loss (MCB-FL). Despite
using Focal Loss~\cite{DBLP:conf/iccv/LinGGHD17} may improve
the accuracy of binary classification in some researches, hand parsing 
is alternatively a multi-class classification task. The parameter
$\alpha_t$ used to balance the positive and negative samples in Focal Loss becomes ineffective. 
Therefore, we generalize Focal Loss, proposing Multi-Class Balanced 
Focal Loss to balance the weights of different classes.

Our MSDB-FCN achieves the state-of-the-art performance on hand parsing
benchmark RHD-PARSING. Our work can make impacts to the computer vision community
since some of our techniques and structures can also be embedded in other
networks. Our main contributions are:
\begin{itemize}
   \item We propose a novel Multi-Scale Dual-Branch Fully Convolutional
   Network (MSDB-FCN) for hand parsing tasks. Since MSDB-FCN employs a Dual-Branch
   architecture to effectively extract the features of a hand region, our network 
   can hence focus on hand parsing rather than processing irrelevant information.

   \item We design a new structure, namely Deconvolution and Bilinear Interpolation
   Block (DB-Block), for upsampling and context encoding. 
   Compared to either applying deconvolution and bilinear interpolation alone,
   the network using DB-Block can better extract and merge multi-scale information.

   \item For tackling data imbalance in multi-class classification, such as image
   parsing and semantic segmentation, we propose Multi-Class Balanced Focal Loss to
   balance the weights of different classes.

\end{itemize}

%%%%%%%%%%%%%%%%%%%%% network architecture
\begin{figure*}
\begin{center}
   \includegraphics[width=1\linewidth]{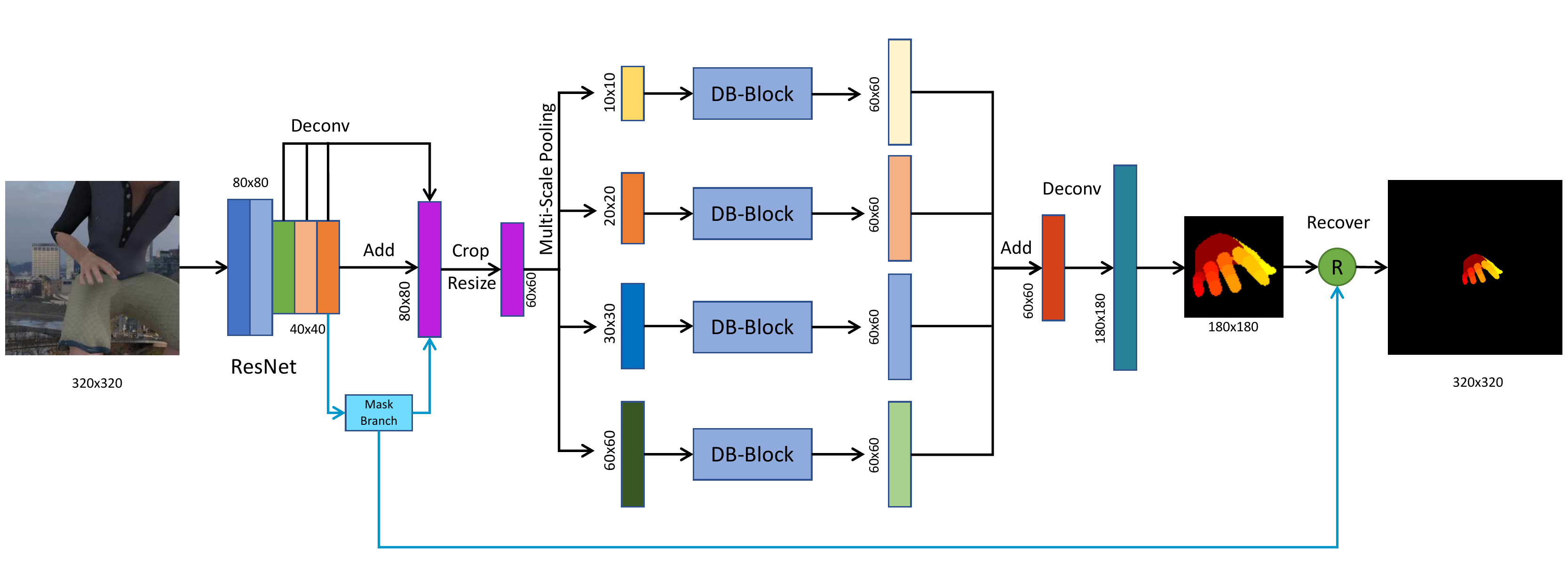}
\end{center}
   \caption{Overview of our MSDB-FCN architecture. With an input image,
   MSDB-FCN uses a ResNet-based model to generate feature maps of different stages.
   We include a mask branch, which is a small subnet, to calculate the hand position
   from the feature map of the last stage. The parsing branch aggregates the output
   of the last three stages to obtain a finer feature map, following by extracting
   hand features with the help of the location information computed by the mask branch.
   After multi-scale pooling and feature fusion, we get a final feature map
   aggregating all information of the hand. Finally, the parsing result of
   hand area is restored to as large as the input image.}
\label{fig:network}
\end{figure*}

%%%%%%%%% Related Work
\section{Related Work}
Our approach is closely related to the areas of (1) object extraction,
(2) multi-scale features aggregation and
(3) loss functions for data imbalance.

{\bf Object Extraction:}
Driven by the development of Convolutional Neural Networks (CNNs),
researches on many computer vision tasks, such as object detection,
semantic segmentation and pose estimation, have achieved remarkable success.
For some specific tasks, object extraction may be required. Methods of
~\cite{DBLP:conf/bmvc/LiAT17,DBLP:conf/eccv/LiL18} combined object detection
network for accurate segmentation. In~\cite{DBLP:conf/iccv/ZimmermannB17}, a
network for segmentation was employed to extract the region of a hand. Yet
these approaches did not make full use of the information extracted by the
network, causing a large number of redundant calculations.
Our MSDB-FCN employs a mask branch to calculate the position of a hand area,
providing prior information to the parsing branch. The mask branch is a small
subnet embedded in the network, which avoids repeated computation of feature
extraction. Compared with the aforementioned methods, the structure of our
model is more complete.

{\bf Multi-Scale Features Aggregation:}
Multi-scale features have been shown to significantly improve the
performance of networks to perform many computer vision tasks. 
A number of researches recently focus on investigating methods
to obtain and encode multi-scale features. 
One way is to use an encoder-decoder structure. 
Chen $et\ al.$~\cite{DBLP:conf/cvpr/ChenWPZYS18} used
a GlobalNet to generate multi-scale features and a RefineNet to encode them.
Lin $et\ al.$~\cite{DBLP:conf/eccv/LinJLCH18} proposed a method of feature
fusion in pairs. Lin $et\ al.$~\cite{DBLP:conf/cvpr/LinMSR17}
designed a multi-path RefineNet network to generate 
high-resolution semantic feature maps. Another way is to employ
multi-scale pooling strategies, such as pyramid pooling.
By controlling the pool size, we can gain features of different scales.
He $et\ al.$~\cite{DBLP:conf/eccv/HeZR014} 
proposed spatial pyramid pooling for visual recognition.
Zhao $et\ al.$~\cite{DBLP:conf/cvpr/ZhaoSQWJ17} extended the
approach and designed PSPNet for scene parsing task.
Yang $et\ al.$~\cite{DBLP:conf/aaai/YangLCGCL18} adjusted the
pool size and designed a bidirectional structure to aggregate
multi-scale features. 
Our work considers the difference of hand parsing from scene parsing.
We then choose suitable pooling sizes and design a novel structure
for upsampling and feature coding. Our model achieves better performance
than using either deconvolution or bilinear interpolation alone.

{\bf Loss Function for Data Imbalance: }
Problems of data imbalance are often encountered in
computer vision tasks. In order to balance different samples,
Shrivastava $et\ al.$~\cite{DBLP:conf/cvpr/ShrivastavaGG16} 
proposed Online Hard Example Mining (OHEM), but it only handled
on easy samples. Class Balanced Loss solved class imbalance
by calculating the loss of positive and negative samples separately
and using weights to balanced them. Focal Loss proposed by
He $et\ al.$~\cite{DBLP:conf/iccv/LinGGHD17} has been shown
great success in object detection, which balances the weights
of different classes and samples. However, Focal Loss is only
effective in binary classification. In multi-class
classification, such as image parsing or semantic segmentation, 
the parameter $\alpha_t$ used in Focal Loss to solving class
imbalance becomes ineffective. In this paper, we propose a
Multi-Class Balanced Focal Loss to enable multi-class
classification, overcoming the limitation of Focal Loss.

%%%%%%%%% Analysis of hand parsing tasks
\section{Analysis of Hand Parsing Tasks}
Comparing with generic object parsing tasks, hand parsing is more challenging
due to its smaller size, complex structure, ambiguous textures and
a large number of self-occlusion. In this section, we use human parsing
as a comparison and further analyze the characteristics and
challenges of hand parsing tasks.

\begin{figure}[t]
\begin{center}
\subfigure[RHD-PARSING]{
   \begin{minipage}[b]{0.22\textwidth}
      \includegraphics[width=\linewidth]{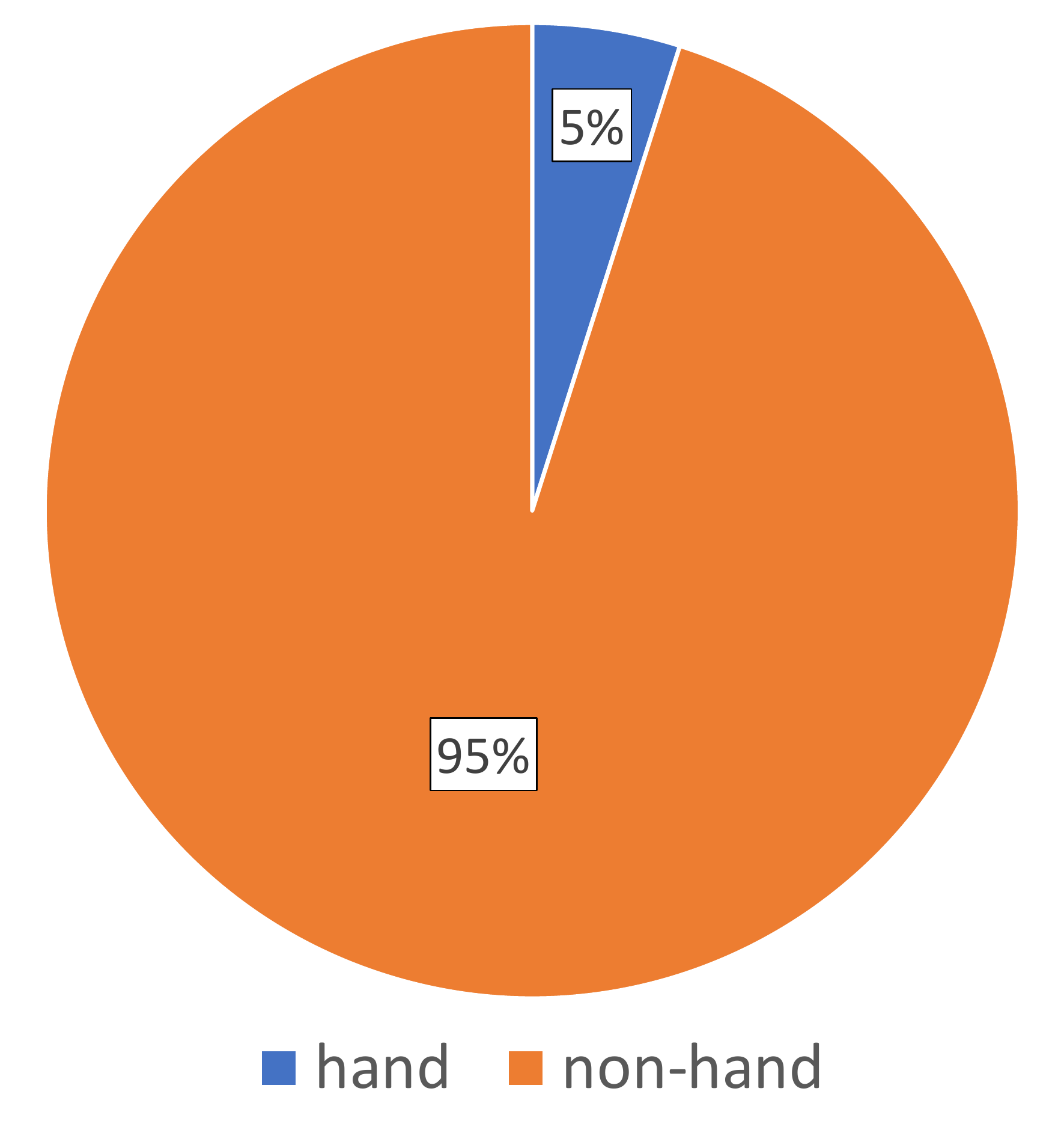}
   \end{minipage}
   }
\subfigure[PASCAL-Person-Part]{
   \begin{minipage}[b]{0.22\textwidth}
      \includegraphics[width=\linewidth]{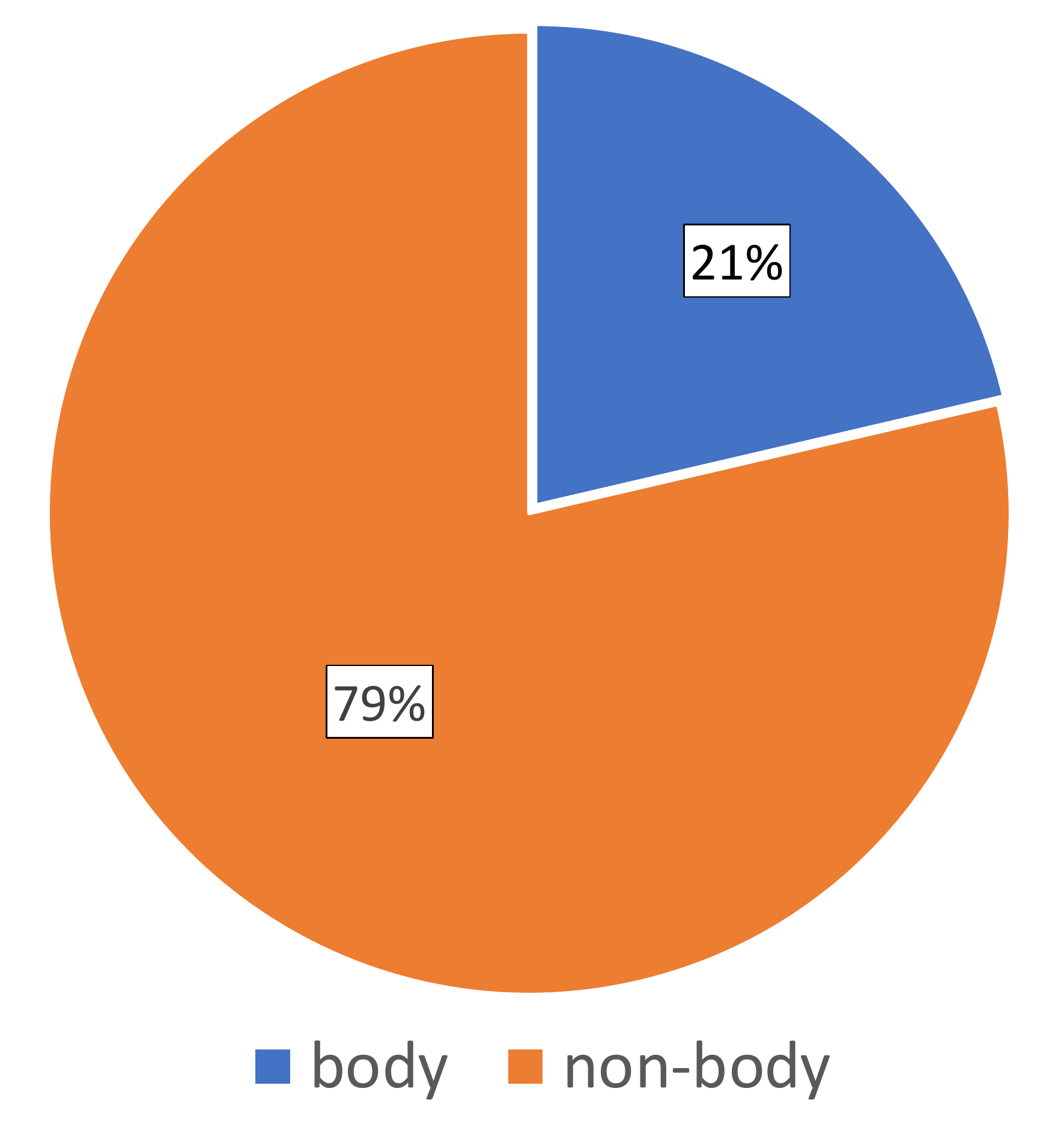}
   \end{minipage}
   } 
\end{center}
   \caption{Comparing the proportion of different objects in images.}
   \label{fig: ratio}
\end{figure}

\begin{figure}[t]
\begin{center}
   \includegraphics[width=\linewidth]{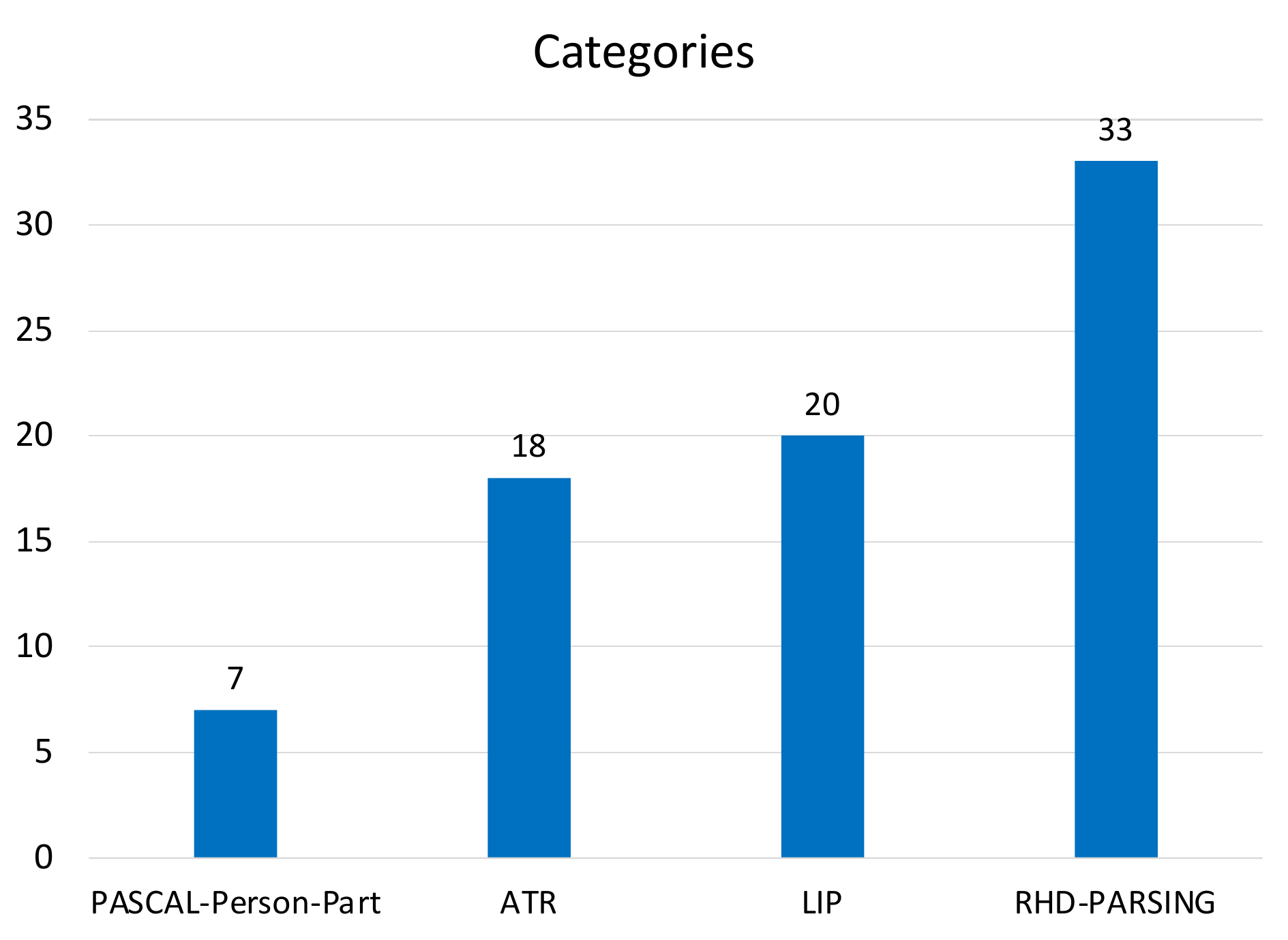}
\end{center}
   \caption{Number of categories in three human parsing datasets and RHD-PARSING.}
   \label{fig: body_hand}
\end{figure}

\begin{figure}[t]
\begin{center}
   \includegraphics[width=\linewidth]{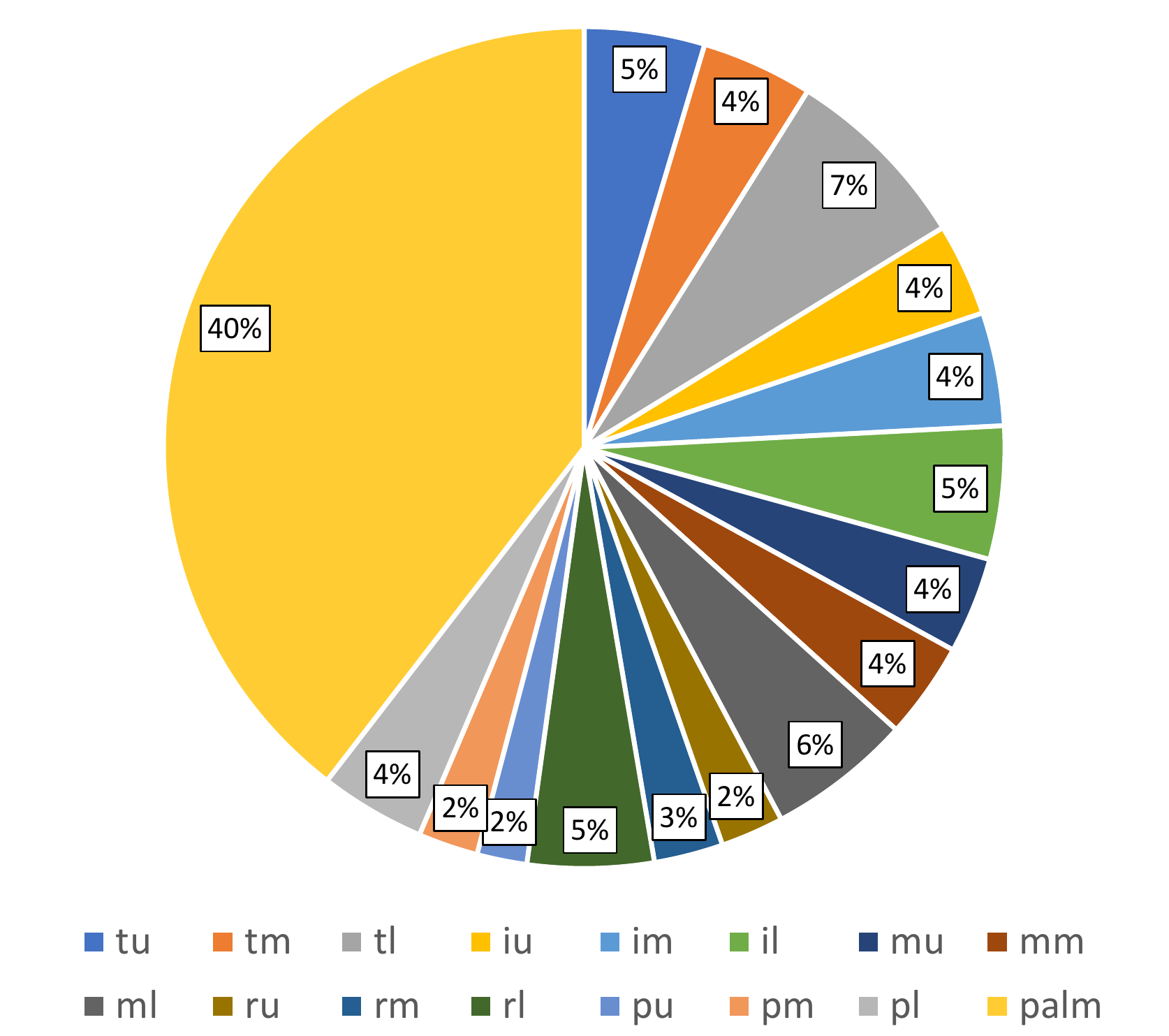}
\end{center}
   \caption{The proportion between diverse parts of a hand. The first letters, 't', 'i',
   'm', 'r' and 'p', of the class names refer to thumb, index finger, middle finger, ring finger    and pinky finger, respectively. The second letters, 'u', 'm' and 'l', denote the different parts of a finger, including the upper, middle and lower.}
   \label{fig: hand parts}
\end{figure}

We analyze the statistics of RHD-PARSING and some human parsing datasets.
As in Figure \ref{fig: ratio}, comparing with human body, hand generally contributes
a much smaller portion of an image. Hence, it is easy to be ignored by a network.
Figure \ref{fig: body_hand} shows the number of segment categories required in existing
human body (PASCAL-Person-Part, ATR, and LIP) and hand (RHD-PARSING) parsing datasets.
In general, hand parsing requires more categories to be segmented, which means
each hand part is extremely small in size. These two characteristics, namely small in part
size and large in number of categories, make accurate hand parsing difficult to achieve.
Figure \ref{fig: hand parts} depicts the distribution of different hand parts in 
RHD-PARSING, indicating the proportion of some parts are imbalanced. Furthermore,
as textures of the whole hand are very similar, it is typically difficult to
distinguish the boundaries between different parts. Also, difficulty of parsing
escalates further due to the existence of a large number of self-occlusion of hand.

A successful algorithm for hand parsing tasks must take into account some hand features.
That is why generic image parsing and semantic segmentation methods cannot parse hand
effectively, as they put excessive attention on the whole image instead, resulting in
extracting significant amount of irrelevant information, such as the background.
Our MSDB-FCN employs a novel architecture to perform well on hand parsing tasks,
which will be introduced in details in Sec. \ref{Sec: MSDB-FCN}.

%%%%%%%%% Multi-Scale Dual-Branch Fully Convolutional Networck
\section{Multi-Scale Dual-Branch Fully Convolutional Network}\label{Sec: MSDB-FCN}
As depicted in Figure \ref{fig:network}, our proposed MSDB-FCN employs a
Dual-Branch structure to better extract the features of hand area. With such
a novel structure, our model can therefore pay more attention on the hand
itself. We also incorporate a new structure, namely Deconvolution and
Bilinear Interpolation Block (DB-Block), for upsampling and feature coding,
gaining excellent results. In addition, we generalize Focal Loss to a
Multi-Class Balanced Focal Loss for performing multi-class classification
tasks. With all of the above, our method achieves the start-of-the-art
performance for hand parsing. We describe our Dual-Branch structure,
DB-Block and Multi-Class Balanced Focal Loss with details in Sec. \ref{Sec: Dual-Branch},
\ref{Sec: DB-Block} and \ref{Sec: MCB-FL}, respectively.

%-------------------------------------------------------------------------
\subsection{Dual-Branch Structure}\label{Sec: Dual-Branch}
\begin{figure}[t]
\begin{center}
   \includegraphics[width=1\linewidth]{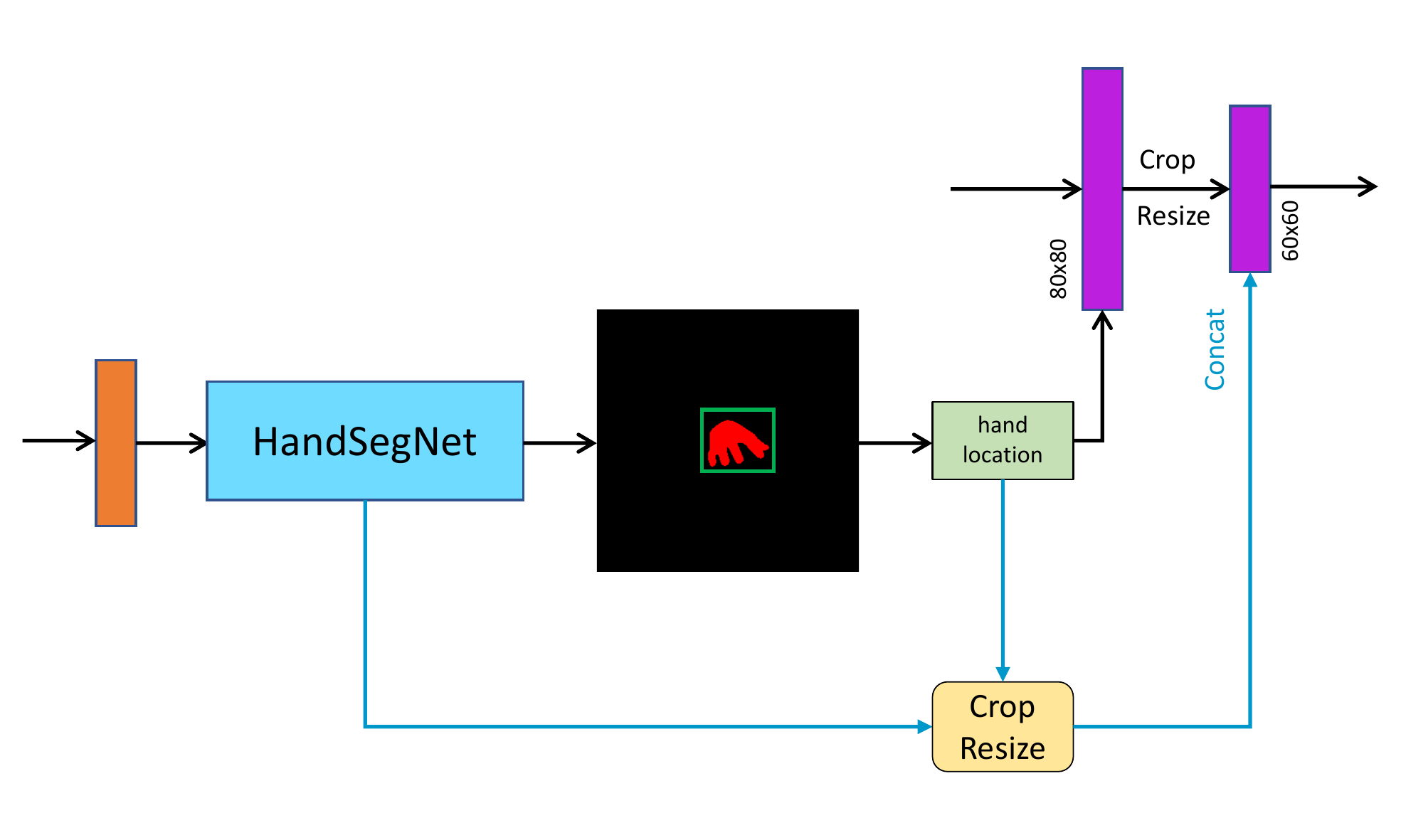}
\end{center}
   \caption{Illustration of the mask branch architecture in our MSDB-FCN.
   Taking feature maps from backbone as input, the location of hand is calculated
   via a small segmentation sub-network and delivered to the parsing branch. The
   mask branch also provides prior information to the parsing branch by feature fusion.}
\label{fig:mask branch}
\end{figure}
As noted previously, scene parsing pays attention to all the components of an
image, while object parsing, like hand parsing, focuses more on the object itself.
For object parsing, background information is not helpful and may even induce
negative impact to the parsing results. For instance, some of the latest methods
for image parsing do not perform well on hand parsing tasks, because a large
number of background features is extracted, which confuses the results. Therefore,
we design a Dual-Branch structure to locate the position of hand in feature maps
and all of the subsequent operations are performed around this area. This allows
our network obtaining more useful information for representing a hand.

The mask branch is a subnet for segmenting the hand, and by using the same features
from a pre-trained backbone model, the parameters of the whole network will not be
increased too much. The architecture of mask branch is illustrated in Figure \ref{fig:mask branch}.
In addition, our mask branch also provides prior knowledge, such as background 
or hand, left or right, to the parsing branch by feature fusion. The parsing
branch aggregates features from different stages of the backbone model to generate
bigger and finer feature maps. With the position information from the mask branch,
the features of hand area are cropped and resized to a fixed size. In order to
gain more contextual information, we employ a multi-scale pooling strategy, which is same as
that in~\cite{DBLP:conf/aaai/YangLCGCL18}, to get multi-scale features. We
denote the features as $f_{s_{i}}\in \{f_{s_{1}},f_{s_{2}}...f_{s_{N}}\}$,
and they are encoded in the following structures. After getting the parsing result
of hand area, the parsing branch reconstructs it into the original image size.

\subsection{Deconvolution and Bilinear Interpolation Block}\label{Sec: DB-Block}

\begin{figure}[t]
\begin{center}
   \includegraphics[width=1\linewidth]{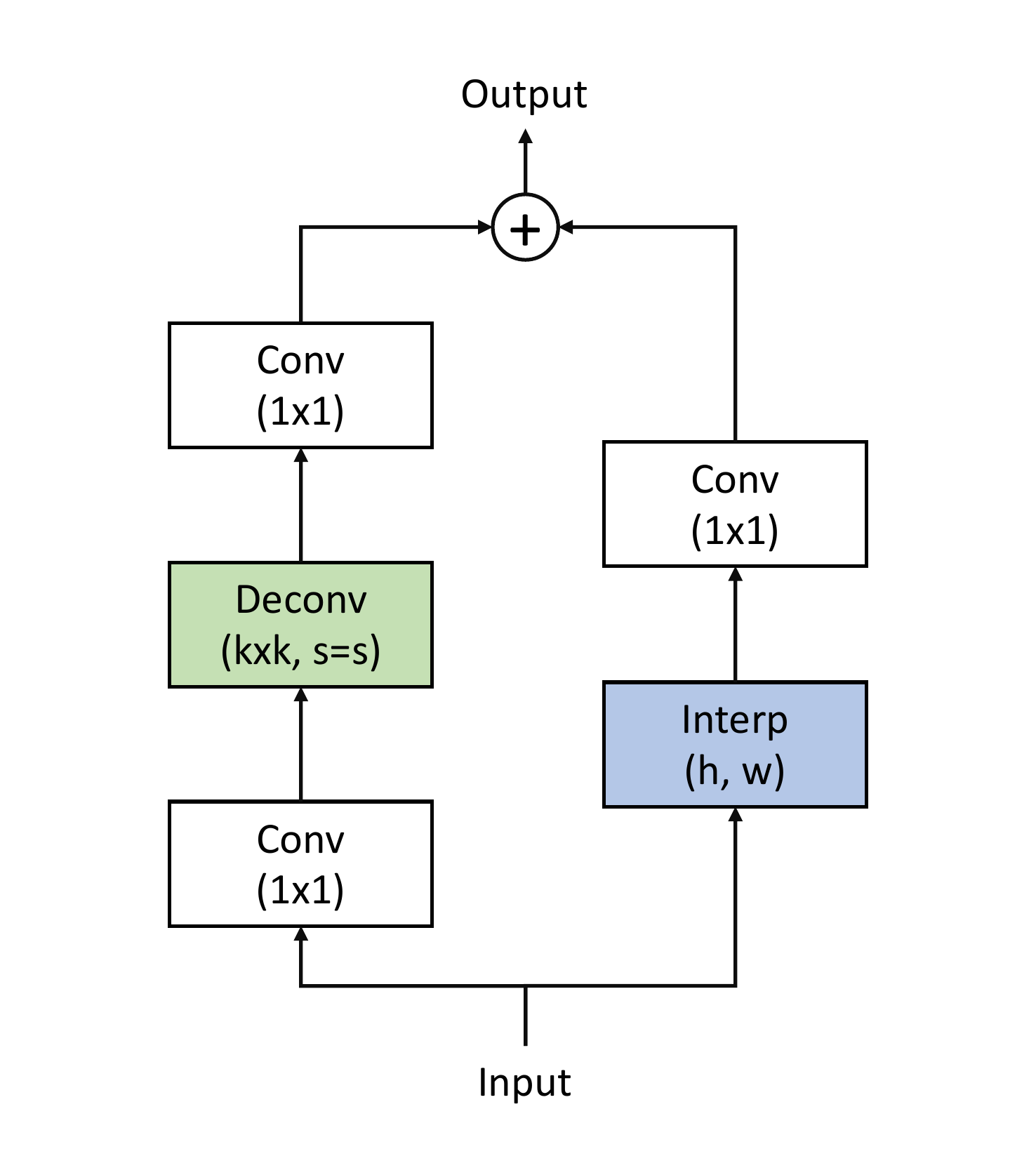}
\end{center}
   \caption{Structure of DB-Block in our MSDB-FCN.}
\label{fig:db block}
\end{figure}

Figure \ref{fig:db block} shows the structure of our Deconvolution and
Bilinear Interpolation Block (DB-Block), which is used to perform upsampling
and feature coding. In many existing works, deconvolution and bilinear interpolation are
the mostly used upsampling methods. The advantage of deconvolution is that the parameters
can be learned, restoring more image details. In contrast, with bilinear interpolation,
intermediate values can be generated by the results of the permutation of surrounding values, while some lost information will not be recovered.
Therefore, when processing from low to high resolution images, main information of the image can be better preserved. 
Inspired by~\cite{DBLP:journals/corr/abs-1812-01187}, 
we design a structure to combine the advantages of deconvolution and bilinear
interpolation. With a residual connection, the features from the deconvolution
layer and the interpolation layer are weighted together. The feature map of the $i$-th
scale can be computed below:
\begin{equation}\label{Eq.1}
h^{f_c}_{s_i}=\alpha_i R^d_{s_i}(f_{s_i})+\beta_i R^b_{s_i}(f_{s_i}),
i<=N 
\end{equation}
where $h^{f_c}_{s_i}$ denotes the feature maps generated by the DB-Block, 
$\alpha_i$ ans $\beta_i$ represent the combination weights.
$R^d_{s_i}(\cdot)$ and $R^b_{s_i}(\cdot)$ denote the resizing feature maps
by deconvolution and bilinear interpolation, respectively.

The updated feature maps are then merged together to generate the final feature
representation $M$ as follows:
\begin{equation}\label{Eq.2}
M=\sum_{i=1}^n\gamma_i h^{f_c}_{s_i}
\end{equation}
where $\gamma_{i}$ is the weight of the $i$-th scale. By combining Eq. \ref{Eq.1}
and Eq. \ref{Eq.2}, the above formula becomes:
\begin{equation}\label{Eq.3}
M=\sum_{i=1}^n \tilde{\alpha_i} R^d_{s_i}(f_{s_i})+\tilde{\beta_i} R^b_{s_i}(f_{s_i})
\end{equation}
where $\tilde{\alpha_i}$ and $\tilde{\beta_i}$ are the products of $\gamma_i$ multiplying
$\alpha_i$ and $\beta_i$, respectively. All weighting processes are
implemented by using $1 \times 1$ convolution layer, whose parameters can be learned.

\subsection{Multi-Class Balanced Focal Loss}\label{Sec: MCB-FL}
Data imbalance, such as class imbalance and sample imbalance,
is a common problem in many computer vision tasks. Class imbalance refers to a large
gap between the number of samples of different classes. Sample imbalance
means that different samples have different classification difficulties. In hand 
parsing tasks, data imbalance is obvious, constituting of the imbalance between
background and hand as well as the imbalance between different hand parts.

Focal Loss~\cite{DBLP:conf/iccv/LinGGHD17} has shown
significant success in binary classification, which can be expressed as:
\begin{equation}\label{Eq.4}
FL(p_t)=-\alpha_t(1-p_t)^\gamma log(p_t)
\end{equation}
where $\alpha_t$ is used to balance positive and negative classes, where the weight of positive and negative classes are $\alpha_t$ and $1-\alpha_t$, respectively.
And the role of $\gamma$ is to balance the samples with different difficulties.
However, when dealing with multi-class classification, having only one parameter $\alpha_t$
becomes insufficient, because it is incapable to give a unique weight for each
different class.

\begin{figure}[t]
\begin{center}
   \includegraphics[width=1\linewidth]{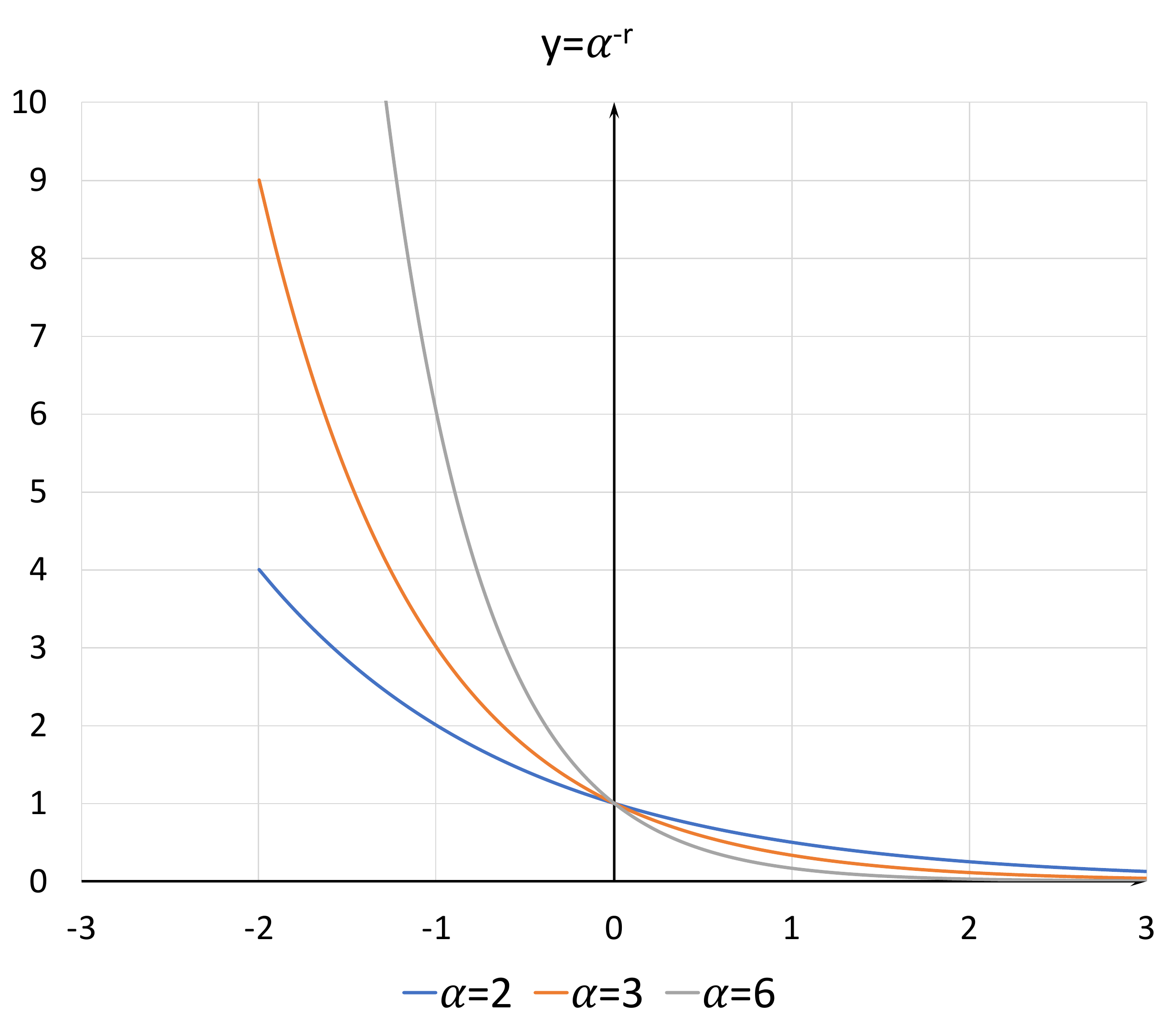}
\end{center}
   \caption{Illustration of the curve of $\alpha^{-r}$ with different values
   of $\alpha$. Parameter $r\in [0, 1]$ represents the proportion of diverse classes
   in an image. Hence, the weights of different classes are limited to $[\alpha^{-1}, 1]$,
   balancing distinct classes with different sample sizes.}
\label{fig:MCB-FL}
\end{figure}

To cope with data imbalance in multi-class classification, we propose a
Multi-Class Balanced Focal Loss. It extends Focal Loss to cover various
semantic segmentation related tasks. The definition of Multi-Class Balanced
Focal Loss is as follows:
\begin{equation}\label{Eq. MCB-FL}
MCB-FL_{all}=\frac{1}{N}\sum_{i=1}^N\sum_{j=1}^C-\alpha^{-r_j}\hat{p}_{ij}(1-p_{ij})^\gamma log(p_{ij})
\end{equation}
where $p_{ij}$ refers to the 
probability of the $i$-th pixel belonging to the class $j$, and $\hat{p}_{ij}$ is the probability 
obtained from the ground truth. $r_j\in [0, 1]$ represents the percentage of pixels belonging to
class $j$ in the image, which can be obtained easily as follows:
\begin{equation}\label{Eq: rj}
r_j=\frac{t_j}{\sum_{k=1}^C t_k}
\end{equation}
where $t_j$ is the total number of pixels of class $j$ in the image. After adding a factor
$r_j$, the parameter $\alpha_t$ in Focal Loss becomes an exponential function $\alpha^{-r_j}$,
where $\alpha$ is a positive real constant. 
As shown in Figure \ref{fig:MCB-FL}, classes with larger proportion have smaller weights, and in
contrast, the weights of classes with smaller proportion are increased. Moreover,
by changing parameter $\alpha$, we can get diverse weighting curves.

%%%%%%%%%%%%%%% experiments
\section{Experiments}
In this section, we depict the dataset and evaluation metrics used, and describe the 
implementation details of our MSDB-FCN. We then report the results of our model and
the comparisons with other state-of-the-art image parsing and semantic segmentation methods.

\subsection{Dataset and Evaluation Metrics}
\noindent
{\bf Dataset }
We evaluate our MSDB-FCN on the benchmark dataset for hand parsing: RHD-PARSING. We merge
human body into the background in the ground truth of the original dataset RHD
~\cite{DBLP:conf/iccv/ZimmermannB17} and create this new dataset with 32 segmentation
masks of hand together with the background for hand parsing tasks. In addition, we merge
all parts of left hand and all parts of right hand to obtain segmentation labels with 3 classes.
The dataset contains 41258 training images and 2728 testing images, comprising images, parsing labels and segmentation labels.\\

\noindent
{\bf Evaluation Metrics }
We use mean pixel accuracy and mean region Intersection over Union as our evaluation metrics.
\begin{itemize}
   \item Mean Accuracy: $\frac{1}{C}\sum_i\frac{n_{ii}}{t_i}$,
   \item Mean IoU: $\frac{1}{C}\sum_i \frac{n_{ii}}{t_i+\sum_j n_{ji}-n_{ii}}$,
\end{itemize}
where $n_{ji}$ means the number of pixels of class $j$ which are predicted to be class $i$, 
$C$ represents the number of parsing classes and $t_i=\sum_j n_{ij}$ is the total number of
pixels of class $i$.

\subsection{Implementation}
\noindent
{\bf Network Architecture }
We implement our approach on platform Keras with tensorflow as the backend.
We use ResNet-50 as proposed by He $et\ al.$~\cite{DBLP:conf/cvpr/HeZRS16} 
to be the pre-trained model. To ensure the last three stages
of ResNet-50 have the same resolution, we adopt the dilated convolution strategy~\cite{DBLP:journals/corr/YuK15}.
Our mask branch is a segmentation subnet with 5 convolution layers, 1 transposed convolution
layer and 1 upsampling layer. The result of segmentation is used to calculate the
location of hand, facilitating the extraction of features of hand region.
For generating multi-scale features, we employ the same pool sizes of pyramid pooling
as in~\cite{DBLP:conf/aaai/YangLCGCL18} for our parsing branch. 
After feature fusion based on DB-Block, the multi-scale features
are encoded together to generate the final result.\\

\noindent
{\bf Training Details }
The resolution of the input image is $320\times320$ pixels. The whole training process
of MSDB-FCN is divided into two stages. Firstly, we use the dataset with segmentation
labels to train the mask branch. Subsequently, the dataset with parsing labels is employed 
to train the parsing branch. We choose Adam as the optimizer and the learning rate is 
set to $10^{-3}$. To reduce overfitting, methods for data augmentation are used,
including random brightness, random channel shift and random contrast with probability
of $1/3$, $1/5$ and $1/7$, respectively. All experiments are conducted on a
single Nvidia Titan X Pascal GPU.

\subsection{Results and Comparisons}
We conduct ablation analysis to evaluate the contribution of diverse elements of our
MSDB-FCN, including the Dual-Branch architecture, DB-Block and the Multi-Class Balanced
Focal Loss. We also compare with state-of-the-art approaches of image parsing and
semantic segmentation on RHD-PARSING dataset.\\

\noindent
{\bf Ablation Analysis for Dual-Branch Architecture }
The mask branch helps the network focus on the region of hand, and reduce the influence
of background. The parsing branch uses multi-scale pooling strategy to generate
multi-scale features and employs DB-Block to encode contextual information of different scales.
In order to evaluate the effect of this Dual-Branch architecture, we compare
the results of the model with different settings. As shown in Table \ref{table: Dual-Branch},
with parsing branch, the Mean IoU of our model improves $4.71\%$ and the Mean Accuracy improves $6.56\%$
compared with the baseline. And with the whole Dual-Branch architecture, our MSDB-FCN achieves $5.31\%$ 
improvement of Mean IoU and $7.52\%$ improvement of Mean Accuracy.\\

\begin{table}
	\begin{center}
		\begin{tabular}{|l|c|c|}
			\hline
			Method & Mean IoU(\%) & Mean Acc.(\%)\\
			\hline\hline
			ResNet50-Baseline& 51.04 & 58.98 \\
			ResNet50+PB& 55.75 & 65.54 \\
			ResNet50+PB+MB& {\bf 56.35}& {\bf 66.50} \\
			\hline
		\end{tabular}
	\end{center}
	\caption{Investigation of the effect of our Dual-Branch architecture. We use the ResNet50-based
		FCN with dilated convolution as the baseline model. 'PB' and 'MB' denote the parsing branch and
		the mask branch, respectively.}
	\label{table: Dual-Branch}
\end{table}

\noindent
{\bf Ablation Analysis for DB-Block Architecture}
Our MSDB-FCN employs DB-Block for upsampling and encoding multi-scale contextual information.
By a residual connection and $1\times1$ convolution, the results of deconvolution and 
bilinear interpolation for upsampling are weighted and merged. As shown in Eq. \ref{Eq.3},
the features of different scales are encoded with diverse weights. In order to evaluate
the effect of our DB-Block, we replace it by deconvolution and bilinear interpolation respectively.
As shown in Table \ref{table: DB-Block}, the MSDB-FCN with DB-Block achieves better performance, which
improves $1.13\%$ of Mean IoU and $1.06\%$ of Mean Accuracy on average.\\

\begin{table}
\begin{center}
\begin{tabular}{|l|c|c|}
\hline
Method & Mean IoU(\%) & Mean Acc.(\%)\\
\hline\hline
MSDB-FCN+B& 55.22 & 65.32 \\
MSDB-FCN+D& 55.22 & 65.57 \\
MSDB-FCN+DB& {\bf 56.35}& {\bf 66.50} \\
\hline
\end{tabular}
\end{center}
\caption{Comparisons of MSDB-FCN with different methods for upsampling, include
deconvolution, bilinear interpolation and DB-Block, which are denoted as 'D', 'B' and 'DB',
respectively.}
\label{table: DB-Block}
\end{table}

\noindent
{\bf Ablation Analysis for Multi-Class Balanced Focal Loss }
In many computer vision tasks, data imbalance is a common problem. Although Focal Loss is
good at binary classification, it is ineffective for multi-class classification tasks,
such as image parsing and semantic segmentation, because
the parameter used to balance positive and negative classes becomes
irrelevant. Our Multi-Class
Balanced Focal Loss generalizes Focal Loss to solve data imbalance
in multi-class classification tasks. We compare the results on baseline model with diverse loss
functions, including categorical cross entropy loss, Focal Loss and our Multi-Class Balanced
Focal Loss, which are shown in Table \ref{table: MCB-FL}. Compared with the model using
categorical cross entropy loss, models with Focal Loss and Multi-Class Balanced Focal Loss
have significantly improved the parsing performance. The model with our loss achieves
$1.33\%$ improvement of Mean IoU and $4.23\%$ improvement of Mean Accuracy comparing with model using
Focal Loss. To find the best weighting curve for hand parsing, we experiment with setting $\alpha$ to
different values. As shown in Table \ref{table: MCB-FL-alpha}, in combination with Mean IoU and Mean Accuracy, we find that $\alpha=3$ works the best.\\

\begin{table}
\begin{center}
\begin{tabular}{|l|c|c|}
\hline
Method & Mean IoU(\%) & Mean Acc.(\%)\\
\hline\hline
ResNet50+CE& 51.04 & 58.98 \\
ResNet50+FL& 53.54 & 62.77 \\
ResNet50+MCB-FL& {\bf 54.87} & {\bf 67.00} \\
\hline
\end{tabular}
\end{center}
\caption{Comparisons of the baseline model with different loss, including categorical cross
entropy loss, Focal Loss and Multi-Class Balanced Focal Loss, which are denoted as 'CE', 'FL'
and 'MCB-FL', respectively.}
\label{table: MCB-FL}
\end{table}

\begin{table}
\begin{center}
\begin{tabular}{|l|c|c|}
\hline
Method & Mean IoU(\%) & Mean Acc.(\%)\\
\hline\hline
ResNet50 ($\alpha=1$)& 53.54 & 62.77 \\
ResNet50 ($\alpha=2$)& {\bf 55.01} & 66.10 \\
ResNet50 ($\alpha=3$)& 54.87 & 67.00 \\
ResNet50 ($\alpha=6$)& 53.85 & {\bf 67.63} \\
\hline
\end{tabular}
\end{center}
\caption{Comparisons of the baseline model with different $\alpha$ values of Multi-Class Balanced Focal Loss.}
\label{table: MCB-FL-alpha}
\end{table}

\begin{table}
\begin{center}
\begin{tabular}{|l|c|c|}
\hline
Method & Mean IoU(\%) & Mean Acc.(\%)\\
\hline\hline
FCN-8s~\cite{DBLP:conf/cvpr/LongSD15}& 38.75& 48.74 \\
DeepLab-V3~\cite{DBLP:journals/corr/ChenPSA17}& 38.42& 51.18 \\
DenseASPP~\cite{DBLP:conf/cvpr/YangYZLY18}& 39.83& 54.46 \\
PSPNet~\cite{DBLP:conf/cvpr/ZhaoSQWJ17}& 42.74& 57.71 \\
SegNet~\cite{DBLP:journals/corr/BadrinarayananK15}& 51.77& 62.13 \\
GCN~\cite{DBLP:conf/cvpr/PengZYLS17}& 51.92& 62.72 \\
RefineNet~\cite{DBLP:conf/cvpr/LinMSR17}& 52.60& 64.44 \\
Bey-SegNet~\cite{DBLP:conf/bmvc/KendallBC17}& 54.79& 65.18 \\
\hline\hline
MSDB-FCN& 56.35 & 66.50\\
MSDB-FCN+MCB-FL& {\bf 57.89} & {\bf 70.23}\\
\hline
\end{tabular}
\end{center}
\caption{Quantitative comparisons of the state-of-the-art image parsing and semantic segmentation methods 
on RHD-PARSING dataset. 
% The top three results are highlighted in Red, Green and Blue, respectively.
}
\label{table: Comparisons with other methods}
\end{table}

\noindent
{\bf Comparisons with Other Methods }
\begin{figure*}
\begin{center}
\subfigure[Image]{
   \begin{minipage}[b]{0.1105\textwidth}
      \includegraphics[width=\linewidth]{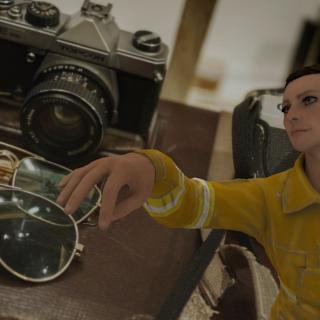}
      \includegraphics[width=\linewidth]{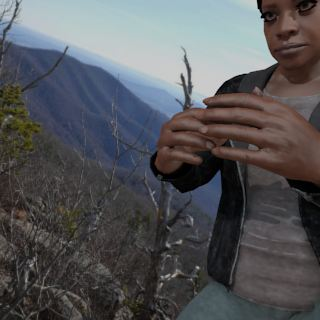}
      \includegraphics[width=\linewidth]{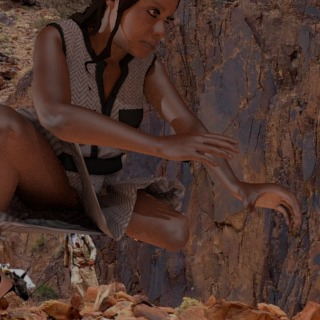}
      \includegraphics[width=\linewidth]{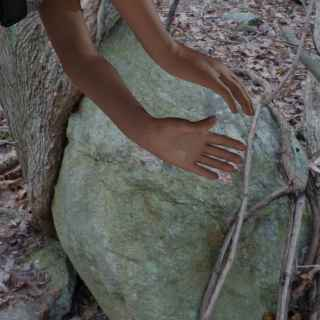}
      \includegraphics[width=\linewidth]{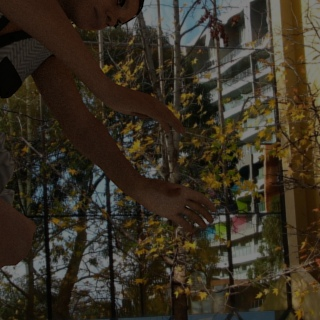}
   \end{minipage}
   }
\subfigure[Ground Truth]{
   \begin{minipage}[b]{0.1105\textwidth}
      \includegraphics[width=\linewidth]{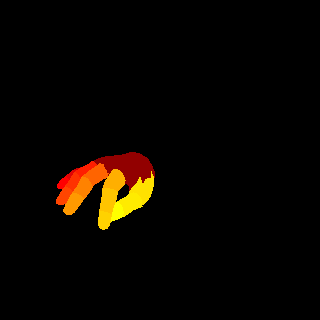}
      \includegraphics[width=\linewidth]{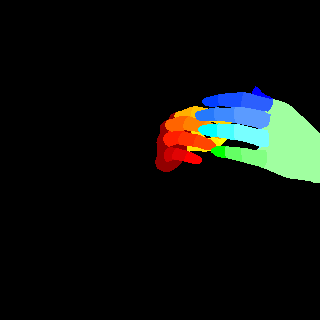}
      \includegraphics[width=\linewidth]{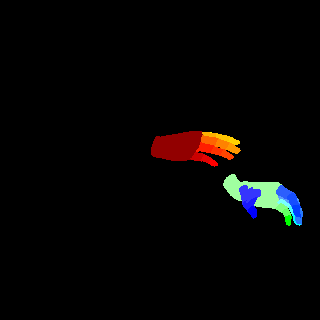}
      \includegraphics[width=\linewidth]{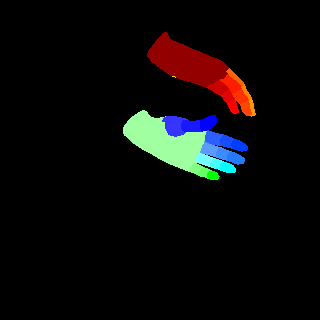}
      \includegraphics[width=\linewidth]{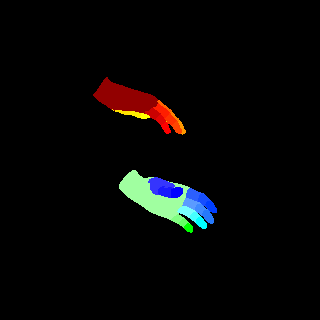}
   \end{minipage}
   } 
\subfigure[FCN-8s]{
   \begin{minipage}[b]{0.1105\textwidth}
      \includegraphics[width=\linewidth]{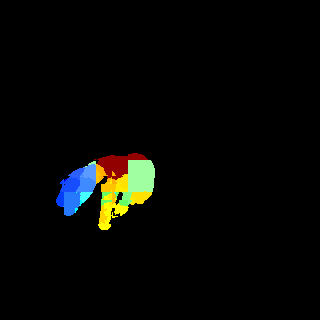}
      \includegraphics[width=\linewidth]{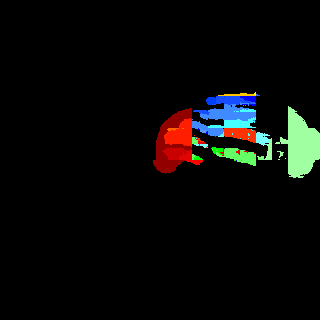}
      \includegraphics[width=\linewidth]{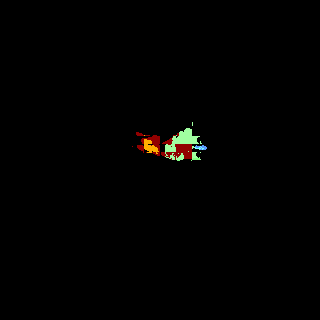}
      \includegraphics[width=\linewidth]{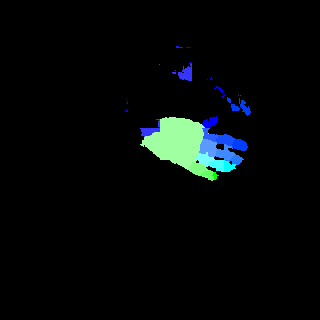}
      \includegraphics[width=\linewidth]{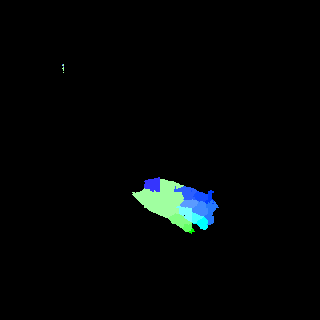}
   \end{minipage}
   } 
\subfigure[DenseASPP]{
   \begin{minipage}[b]{0.1105\textwidth}
      \includegraphics[width=\linewidth]{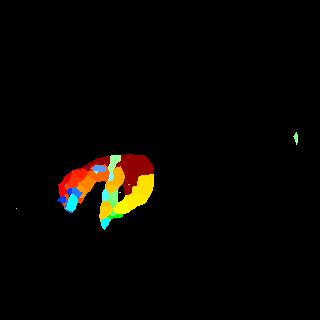}
      \includegraphics[width=\linewidth]{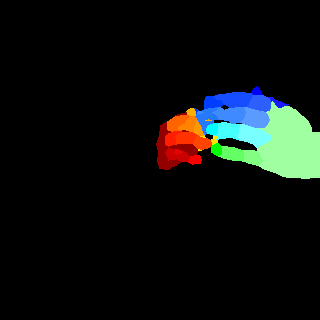}
      \includegraphics[width=\linewidth]{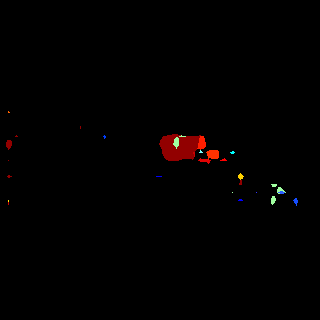}
      \includegraphics[width=\linewidth]{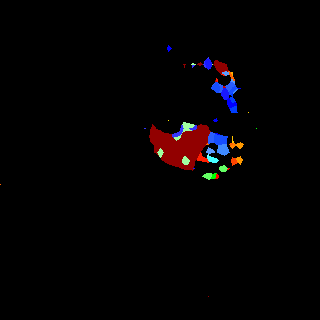}
      \includegraphics[width=\linewidth]{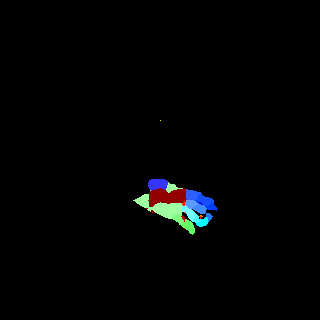}
   \end{minipage}
   } 
\subfigure[PSPNet]{
   \begin{minipage}[b]{0.1105\textwidth}
      \includegraphics[width=\linewidth]{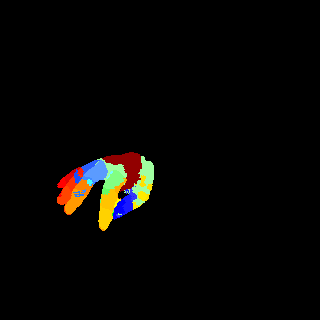}
      \includegraphics[width=\linewidth]{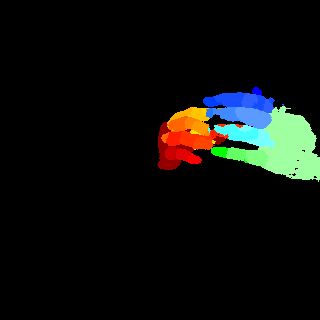}
      \includegraphics[width=\linewidth]{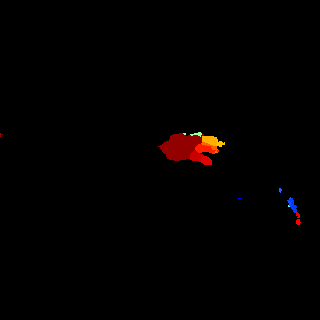}
      \includegraphics[width=\linewidth]{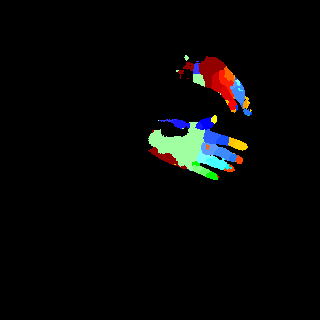}
      \includegraphics[width=\linewidth]{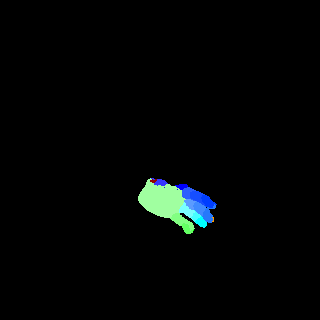}
   \end{minipage}
   } 
\subfigure[RefineNet]{
   \begin{minipage}[b]{0.1105\textwidth}
      \includegraphics[width=\linewidth]{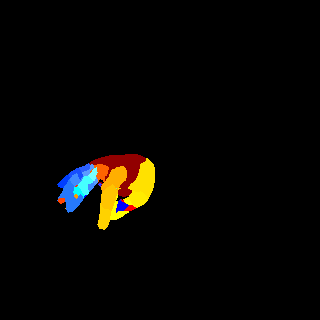}
      \includegraphics[width=\linewidth]{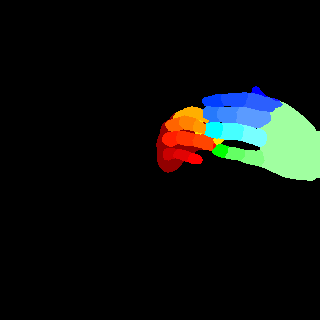}
      \includegraphics[width=\linewidth]{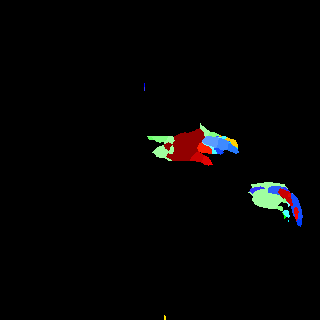}
      \includegraphics[width=\linewidth]{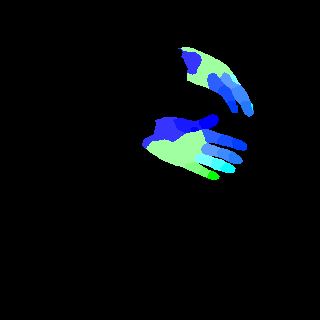}
      \includegraphics[width=\linewidth]{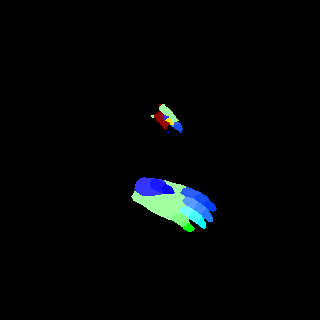}
   \end{minipage}
   } 
   \subfigure[Bey-SegNet]{
   \begin{minipage}[b]{0.1105\textwidth}
      \includegraphics[width=\linewidth]{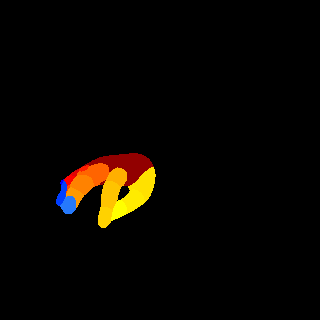}
      \includegraphics[width=\linewidth]{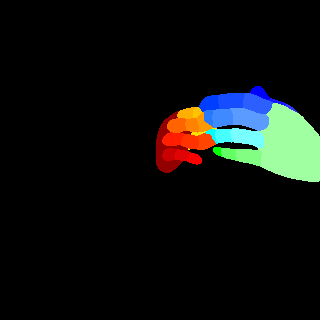}
      \includegraphics[width=\linewidth]{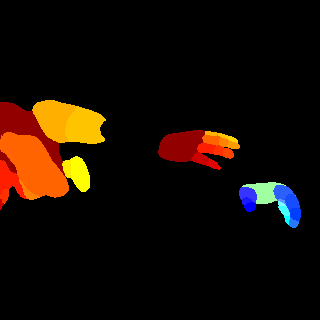}
      \includegraphics[width=\linewidth]{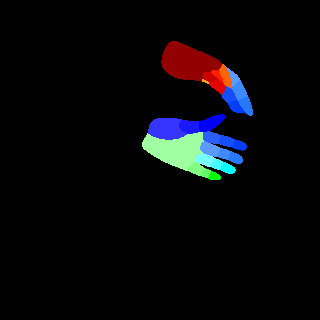}
      \includegraphics[width=\linewidth]{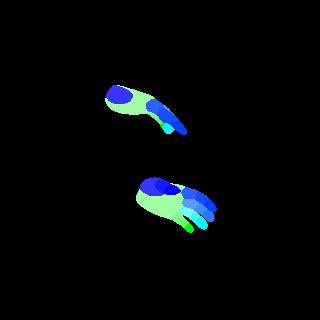}
   \end{minipage}
   } 
\subfigure[MSDB-FCN]{
   \begin{minipage}[b]{0.1105\textwidth}
      \includegraphics[width=\linewidth]{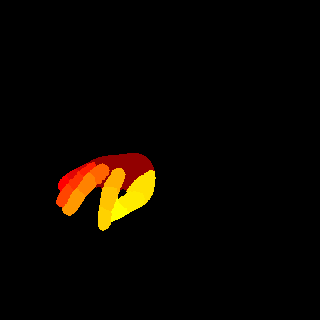}
      \includegraphics[width=\linewidth]{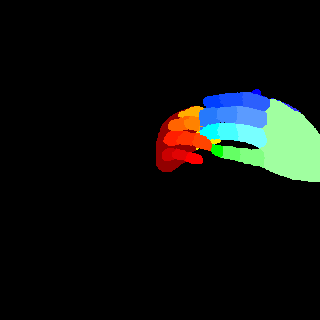}
      \includegraphics[width=\linewidth]{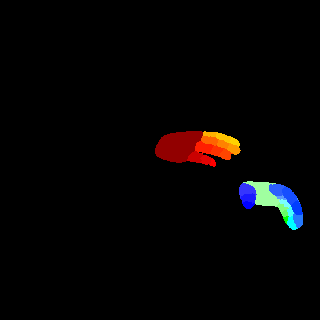}
      \includegraphics[width=\linewidth]{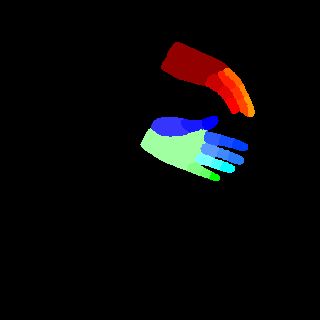}
      \includegraphics[width=\linewidth]{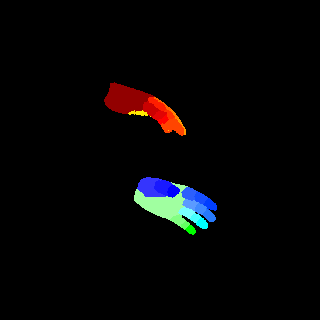}
   \end{minipage}
   } 
\end{center}
   \caption{Example results of existing image parsing and semantic segmentation methods as well as our MSDB-FCN on RHD-PARSING dataset. (a) Selected images from RHD-PARSING dataset. (b) Corresponding parsing labels of RHD-PARSING dataset. 
   (c)-(g) Results from existing image parsing and semantic segmentation methods. (h) Results from our MSDB-FCN.}
   \label{fig: Examples}
\end{figure*}
We compare the proposed method with other leading image parsing and semantic segmentation methods 
on RHD-PARSING dataset. As shown in Table \ref{table: Comparisons with other methods},
our MSDB-FCN is more effective for hand parsing tasks and achieves the state-of-the-art performance. 
However, some approaches which are successful in scene parsing and semantic segmentation do not 
achieve satisfactory results. For further comparison, Figure \ref{fig: Examples} depicts hand parsing results from images, which comprise complex background, heavy self-occlusion and small size of hand, by running existing and our methods. Obviously, existing methods generally perform badly, producing results with incorrect hand part classification, missing hand details and mis-classification of part of the background as hand parts. This is mainly due to the excessive attention to the whole image. Object parsing, especially in small object parsing, 
should pays more attention to the object itself, since the large amount of background information does not contribute anything relevant. Therefore, some strategies, which extract multi-scale context of the whole image, are usually ineffective. In contrast, our method outperforms existing methods and can generate hand parsing results well matching with the ground truths.

%%%%%%%%%% Conclusion
\section{Conclusion}
We have proposed a novel parsing framework, called MSDB-FCN, for hand parsing tasks.
Our MSDB-FCN employs a Dual-Branch architecture, including a mask branch and a parsing branch,
to accurately extract the features of hand region, which makes the network pays
more attention to the hand rather than other background information. To better
encode the multi-scale features, we use DB-Block for upsampling and feature fusion.
The DB-Block combines the advantages of Deconvolution and Bilinear Interpolation and employs 
a residual connection and $1\times1$ convolution to merge the results with different weights. 
On the other hand, data imbalance is a common problem
in many computer vision tasks as well as in hand parsing. We have extended Focal Loss to
propose the Multi-Class Balanced Focal Loss for image parsing and semantic segmentation.
Using such a loss in the network, our MSDB-FCN achieves the state-of-the-art
performance in hand parsing tasks.

{\small
\bibliographystyle{ieee}
\bibliography{egbib}
}

\end{document}